\documentclass[10pt,twocolumn,letterpaper]{article}

\usepackage{cvpr}
\usepackage{times}
\usepackage{epsfig}
\usepackage{graphicx}
\usepackage{amsmath}
\usepackage{amssymb}
\usepackage{multirow}
\usepackage{url}
\usepackage{enumitem}
\usepackage{authblk}
\usepackage{stfloats}

\usepackage[pagebackref=true,breaklinks=true,letterpaper=true,colorlinks,bookmarks=false]{hyperref}

\cvprfinalcopy 

\def\cvprPaperID{2302} 

\ifcvprfinal\fi
\begin{document}

\title{Boosting Semantic Human Matting with Coarse Annotations}


\author{Jinlin Liu$^{1,2}$ \qquad  Yuan Yao$^{1}$ \qquad  Wendi Hou$^1$ \qquad  Miaomiao Cui$^1$ \vspace{-10pt}\\
 Xuansong Xie$^1$  \qquad  Changshui Zhang$^2$ \qquad Xian-sheng Hua$^{1}$ \vspace{3pt}\\
\normalsize$^1$Alibaba Group, \qquad \normalsize$^2$ Department of Automation, Tsinghua University\\
\tt\small \{ljl191782, ryan.yy, wendi.hwd, miaomiao.cmm\}@alibaba-inc.com~~xingtong.xxs@taobao.com \\
\tt\small zcs@mail.tsinghua.edu.cn\qquad xiansheng.hxs@alibaba-inc.com}

%
%
%


\maketitle
\thispagestyle{empty}
\begin{abstract}
Semantic human matting aims to estimate the per-pixel opacity of the foreground human regions. It is quite challenging and usually requires user interactive trimaps and plenty of high quality annotated data. Annotating such kind of data is labor intensive and requires great skills beyond normal users, especially considering the very detailed hair part of humans. In contrast, coarse annotated human dataset is much easier to acquire and collect from the public dataset. In this paper, we propose to use coarse annotated data coupled with fine annotated data to boost end-to-end semantic human matting without trimaps as extra input. Specifically, we train a mask prediction network to estimate the coarse semantic mask using the hybrid data, and then propose a quality unification network to unify the quality of the previous coarse mask outputs. A matting refinement network takes in the unified mask and the input image to predict the final alpha matte. The collected coarse annotated dataset enriches our dataset significantly, allows generating high quality alpha matte for real images. Experimental results show that the proposed method performs comparably against state-of-the-art methods. Moreover, the proposed method can be used for refining coarse annotated public dataset, as well as semantic segmentation methods, which reduces the cost of annotating high quality human data to a great extent.
\end{abstract}

\section{Introduction}

Human matting is an important image editing task which enables accurate separation of humans from their backgrounds. It aims to estimate the per-pixel opacity of the foreground regions, making it valuable to use the extracted human image in some recomposition scenarios, including digital image and video production. One may refer this task as semantic segmentation problem~\cite{badrinarayanan2017segnet,chen2017rethinking,long2015fully}, which achieves fine-grained inference for enclosing objects. However, segmentation techniques focus on pixel-wise binary classification towards scene understanding, although semantic information is well labelled, it could not catch complicated semantic details like human hair.

\begin{figure}[t]
  \centering
  \resizebox{1.00\linewidth}{!}{
   \includegraphics{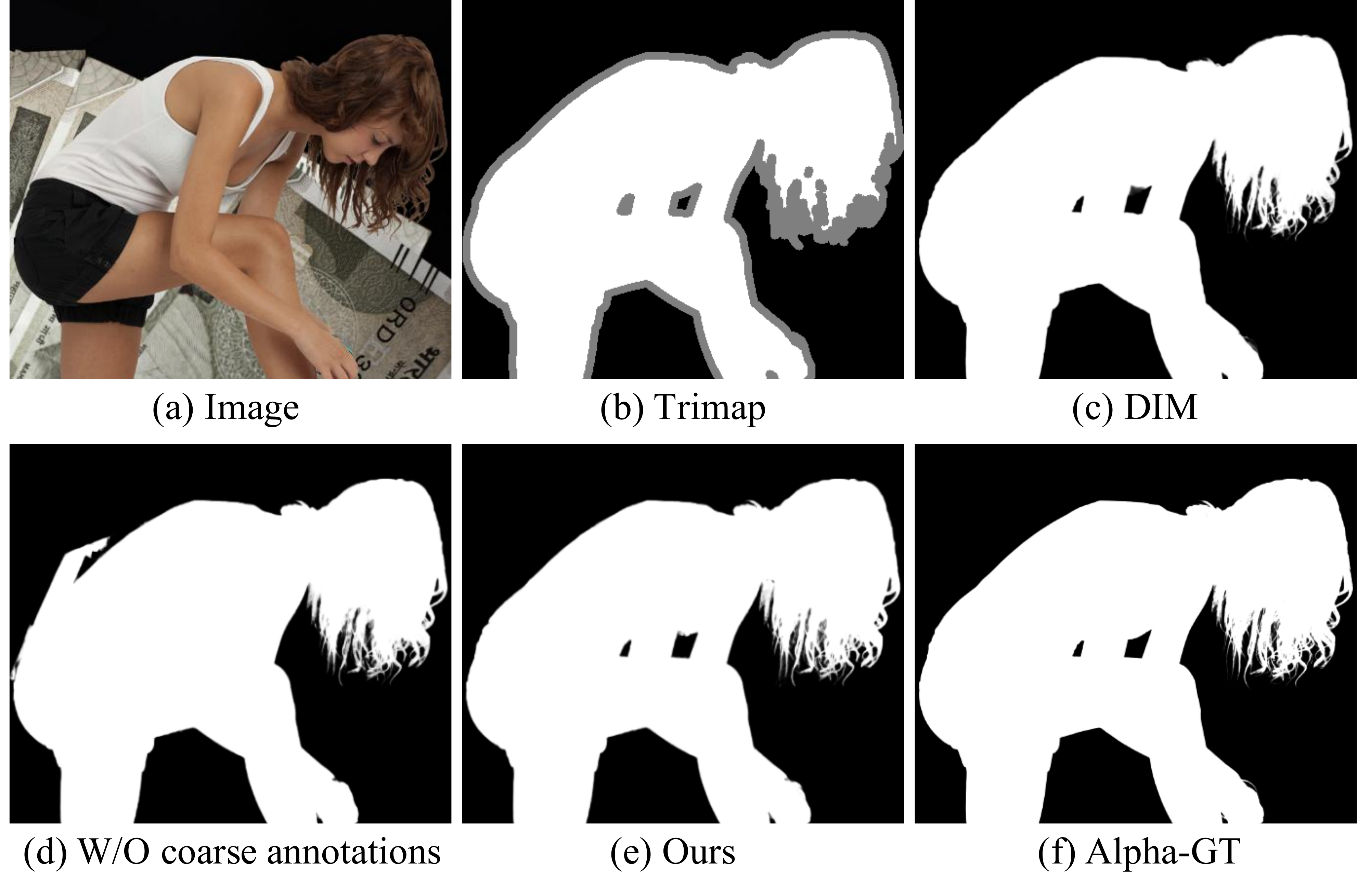} }
  \caption{The user interactive method could catch precise semantics and details under the guidance of trimaps. Without the trimap and enough training dataset, one may get inaccurate semantic estimation, which inevitably leads to wrong matting results. Our methods achieve comparable matting results by leveraging coarse annotated data while do not need trimaps as inputs.}
  \label{fig: motivation}
\end{figure}

The matting problem can be formulated in a general manner. Given an input image $I$, matting is modeled as the weighted combination of foreground image $F$ and background image $B$ as follows~\cite{wang2008image}:
\begin{equation}
\label{eq:matting_def}
I_{z} = \alpha_{z}F_{z} + (1 - \alpha_{z})B_{z}\,, \quad \alpha_{z} \in [0,1]\,.
\end{equation}
where $z$ represents any pixel in image $I$. The known information in Eq.~\ref{eq:matting_def} are the three dimensional RGB color $I_{z}$, while the RGB color $F_{z}$ and $B_{z}$, and the alpha matte estimation $\alpha_{z}$ are unknown. Matting is thus to solve the 7 unknown variables from 3 known values, which is highly under-constrained. Therefore, most existing matting methods take a carefully specified trimap as constraint to reduce the solution space. However, a dilemma in terms of quality and efficiency for trimaps still exists.

The key factor that affecting the performance of matting algorithm is the accuracy of trimap. The trimap divides the image into three regions, including the definite foreground, the definite background and the unknown region. Intuitively, the smaller regions around foreground boundary that the trimap contains, the less unknown variables would be estimated, leading to a more precise alpha matte result. However, designing such an accurate trimap requires a lot of human efforts with low efficiency. The labeling quality should be unified among all the data, either large or small size of unknown regions will degrade the final alpha matte effects. One possible solution to solve the dilemma is adaptively learn a trimap from coarse to fine~\cite{shen2016deep,Cai_2019_ICCV}. In contrast, another solution discards the trimap from the input and employs it as an implicit constraint to a deep matting network\cite{chen2018semantic,Zhang_2019_CVPR}. However, these methods still rely on the quality of the generated trimap, unable to retain both the semantic information and high quality details when implicit trimap is inaccurate.

Another limitation comes from the data for human matting. It is important to have high quality annotation data for image matting task. Since humans in natural images possess a variety of colors, poses, head positions, clothes, accessories, etc. The semantically meaningful structure around the foreground like human hair, furs are the challenging regions for human matting.  Annotating such accurate alpha matte is labor intensive and requires great skills beyond normal users. Shen \etal~\cite{shen2016deep} proposed a human portrait dataset with 2000 images, but it has strict constraint on position of human upper body. The widely used DIM dataset~\cite{xu2017deep} is limited in human data, with only 213 human images. Although Chen \etal~\cite{chen2018semantic} created a large human matting dataset, it is only for commercial use. Unfortunately, collecting the dataset in ~\cite{chen2018semantic} with 35,311 images takes more than 1,200 hours, which is undesirable in practice. Therefore, we argue that there is a solution by combining the limited fine annotated image with easily collected coarse annotated image for human matting.

To address the aforementioned problems, we propose a novel framework to utilize both coarse and fine annotated data for human matting. Our method could predict accurate alpha matte with high quality details and sufficient semantic information without trimap as constraint, as shown in Figure~\ref{fig: motivation}. We achieve this goal by proposing a coupled pipeline with three subnetworks. The mask prediction network (MPN) aims to predict low resolution coarse mask, which contains semantic human information. MPN is trained using both fine and coarse annotated data for better performance on various real images. However, the output of MPN may vary and are not consistent with respect to different input images. Therefore, a quality unification network (QUN) trained on hybrid annotated data is introduced to rectify the quality level of MPN output to the same level. A matting refinement network (MRN) is proposed to predict the final accurate alpha matte, taking in both the origin image and its unified coarse mask as input. Different with MPN and QUN, the matting refinement network is trained using only the fine annotated data.


We also constructed a hybrid annotated dataset for human matting task. The dataset consists of both high quality (fine) annotated human images and low quality (coarse) annotated human images. We first collect 9526 images/alpha pairs with fine annotations. In comparison with previous dataset, we diversity the distribution of human images with carefully annotated alpha matte~\cite{shen2016deep,xu2017deep}, within a labor rational volume size~\cite{chen2018semantic}. We further collect 10597 coarse annotated data to better capture accurate semantics within our framework. We follow~\cite{xu2017deep} to composite both data onto 10 background images in MS COCO~\cite{lin2014microsoft} and Pascal VOC~\cite{everingham2010pascal} to form our dataset. Comprehensive experiments have been conducted on this dataset to demonstrate the effectiveness of our method, and our model is able to refine coarse annotated public dataset as well as semantic segmentation methods, which further verifies the generalization of our method. The main contributions of this work are:

\begin{itemize}[topsep=0.5pt, itemsep=0.5pt, partopsep=0.5pt]
\item To our best knowledge, this is the first method that uses coarse annotated data to enhance the performance of end-to-end human matting. Previous methods either take trimap as constraint or use sufficient fine annotated dataset only.
\item We propose a quality unification network to rectify the mask quality during the training process so as to utilize both coarse and fine annotations, allowing accurate semantic information as well as structural details.
\item The proposed method can be used to refine coarse annotated public dataset as well as semantic segmentation methods, which makes it easy to create fine annotated data from coarse masks.
\end{itemize}







\begin{figure*}[ht]
  \centering
  \resizebox{1\linewidth}{!}{
   \includegraphics{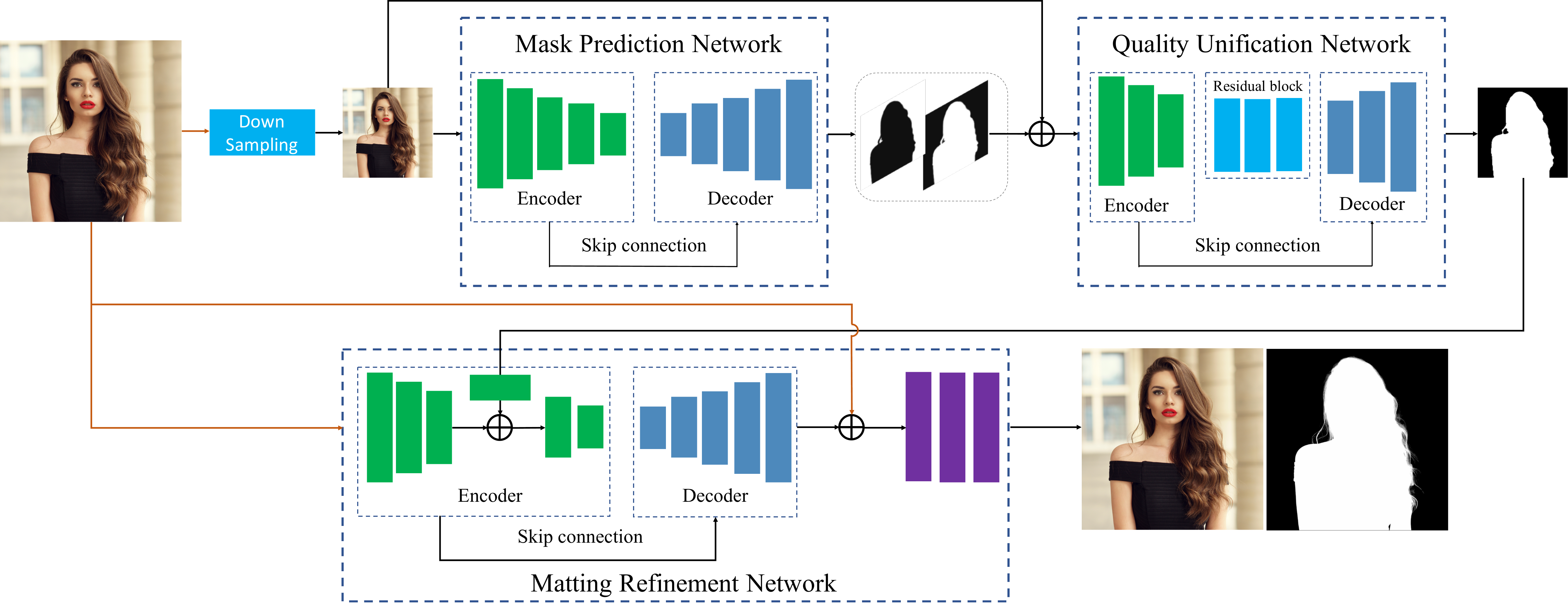} }
  \caption{An overview of our network architecture. The proposed method is composed of three parts. The first part is mask prediction network (MPN), to predict low resolution coarse semantic mask. MPN is trained using both coarse and fine data. The second part is quality unification network (QUN). QUN aims to rectify the quality of the output from the mask prediction network to the same level. The rectified coarse mask is then unified and enables consistent input for training the following alpha matte prediction stage. The third part is matting refinement network (MRN), taking in the input image and the unified coarse mask to predict the final accurate alpha matte.}
  \label{fig: flowchart}
\end{figure*}

\section{Related Work}
\noindent {\bf Natural Image Matting.} Natural image matting tries to estimate the the unknown area with known foreground and background in the trimap.

The traditional methods can be summarized to sampling based methods and affinity based methods~\cite{wang2008image}. The sampling based methods~\cite{chuang2001bayesian,feng2016cluster,gastal2010shared,he2011global,johnson2016sparse,karacan2015image,ruzon2000alpha} leverage the nearby known foreground and background colors to infer the alpha values of the pixels in the undefined area. Assuming that alpha values for two pixels have strong correlations if the corresponding colors are similar. Following the assumption, various sampling methods are proposed including Bayesian matting~\cite{chuang2001bayesian}, sparse coding~\cite{feng2016cluster,johnson2016sparse}, global sampling~\cite{he2011global} and KL-divergence approaches~\cite{karacan2015image}. Compared with sampling based methods, Affinity based methods~\cite{aksoy2018semantic,aksoy2017designing,bai2007geodesic,chen2013knn,grady2005random,levin2007closed,levin2008spectral,sun2004poisson} define different affinities between neighboring pixels, trying to model the matte gradient instead of the per-pixel alpha value.
%

Deep learning based method is able to learn a mapping between the image and corresponding alpha matte in an end-to-end manner. Cho \etal~\cite{cho2016natural} take the advantage of close-form matting~\cite{levin2007closed} and KNN matting~\cite{chen2013knn} for alpha mattes reconstruction. Xu \etal~\cite{xu2017deep} integrate the encoder-decoder structure with a following refinement network to predict alpha matte. Lutz \etal~\cite{lutz2018alphagan} further employ the generative adversarial network for image matting task. Cai \etal~\cite{Cai_2019_ICCV} argue the limitation of directly estimating the alpha matte from a coarse trimap, and propose to disentangle the matting into trimap adaptation and alpha estimation tasks. Compared with the above methods, our method simply use RGB images as input without the constraint of designated trimaps.

\noindent {\bf Human image Matting.} As a specific type of image matting, human matting aims to estimate the accurate alpha matte corresponding to the human in the input image, which involves semantically meaningful structures like hair. Recently, several deep learning based human matting methods~\cite{chen2018semantic,shen2016deep,zhu2017fast} have been proposed. Shen \etal~\cite{shen2016deep} propose a deep neural network to generate the trimap of a portrait image and add a matting layer\cite{levin2007closed} for network optimization using the forward and backward propagation strategy. Zhu \etal~\cite{zhu2017fast} use a similar pipeline and design a light dense network for portrait segmentation and a feature block to learn the guided filter~\cite{he2010guided} for alpha matte prediction. Chen \etal~\cite{chen2018semantic} introduce an automatic human matting algorithm without feeding trimaps. It combines a segmentation module with a matting module for end-to-end matting. The late fusion CNN structure in ~\cite{Zhang_2019_CVPR} integrates the foreground and background classification presents its capacity for human image matting. However, these models require carefully collected image/alpha pairs, which may also suffer from inaccurate semantics due to lack of fine annotated human data.

\section{Proposed Approach}

We develop three subnetworks as a sequential pipeline. The first one is mask prediction network (MPN), to predict coarse semantic masks using data at different annotation quality level. The second one is quality unification network (QUN). QUN rectifies the quality of the output coarse mask from MPN to the same level. The third part is matting refinement network (MRN), to predict the final accurate alpha matte. The flowchart and the network structure is displayed in Figure~\ref{fig: flowchart}.

\subsection{Mask Prediction Network}
As no trimap is required as input, the first stage of the proposed method is to predict a coarse semantic mask. The network we use is encoder-decoder structure with skip connection, and we predict the foreground mask and the background mask at the same time. At this stage, we aim to estimate a coarse mask, and therefore the network is not trained at a high resolution. We resize all training data to resolution $192\times 160$ so as to train the mask prediction network (MPN) efficiently. In addition, the mask prediction network is trained using all training data, including low quality and high quality annotated data. The loss function to train LRPN is $L_1$ loss,
\begin{equation}
\begin{aligned}
\label{eq:l1_loss}
\mathcal{L}_{MPN}=\lambda_L|\alpha^c_p-\alpha^c_g|_1+(1-\lambda_L)|\beta^c_p-\beta_g^c|_1\,,
\end{aligned}
\end{equation}
where the output is a 2-channel mask, $\alpha^c_p$ denotes the first channel of the output, i.e., the predicted foreground mask, $\alpha^c_g$ denotes the ground truth foreground mask, $\beta^c_p$ denotes the second channel of the output, i.e., the predicted background mask, and $\beta_g^c$ denotes the ground truth background mask. We set $\lambda_L=0.5$ in experiments.

\subsection{Quality Unification Network}
\begin{figure}[t]
  \centering
  \resizebox{0.9\linewidth}{!}{
   \includegraphics{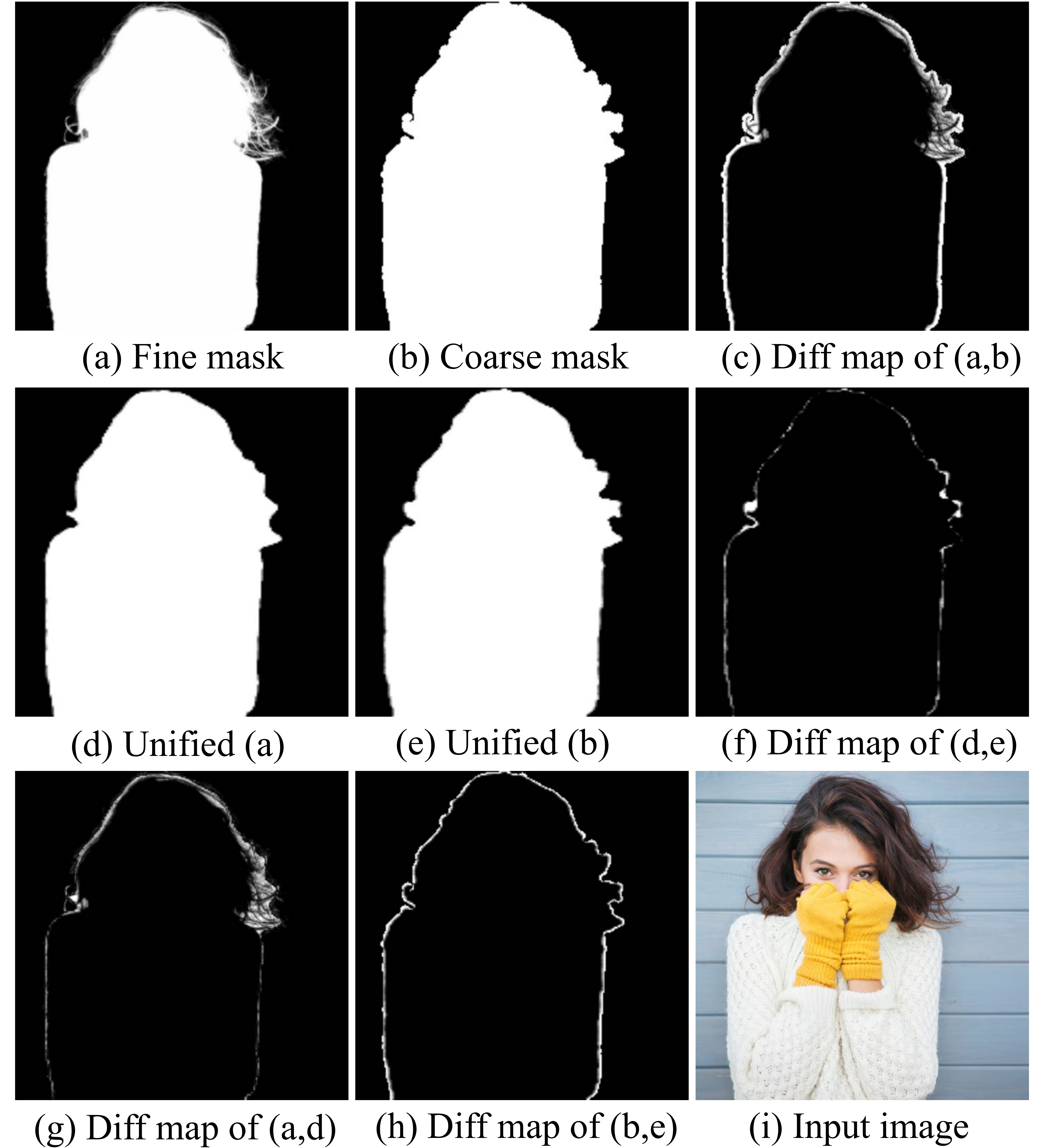} }
  \caption{Different quality of masks are unified by QUN. (a) High quality mask. (b) low quality mask. (c) Difference map of high and low quality mask. (d) Unified result of high quality mask by QUN. (e) Unified result of low quality mask by QUN. (f) Difference map of the unified high quality mask and the low quality mask. (g) Difference map of the unified high quality mask and the original high quality mask. (h) Difference map of the unified low quality mask and the original low quality mask. (i) Input image. }
  \label{fig:demo_mqun}
\end{figure}

Due to the high cost of annotating high quality matting data, we propose to use hybrid data from different data source. Some of the data is annotated at high quality , even hairs are very well separated from the background (Figure~\ref{fig:demo_mqun}(a)). Whereas, majority of other data are annotated at a relatively low quality (Figure~\ref{fig:demo_mqun}(b)). Mask prediction network is trained with both fine annotated data and coarse annotated data. Thus, the quality of the predicted mask may vary significantly. As the alpha matte prediction network can only be trained on the high quality annotated data, the variation of the coarse mask quality will inevitably lead to inconsistent matting results during the inference stage. As illustrated in Figure~\ref{fig:mid_figure}(c), if the coarse mask is relatively accurate, the refinement network will work well to output accurate alpha matte. On the contrary, the refinement network will fail if the coarse mask lacks important details.

We proposed to eliminate the data bias for training matting refinement network by introducing a quality unification network (QUN). The quality unification network aims to rectify the output quality of the mask prediction network to the same level, by improving the quality of coarse masks and lowering the quality of fine masks simultaneously. The output of the mask prediction network and the original image are feed into the quality unification network to unify the quality level. The rectified coarse mask is unified and enables consistent input for training the following accurate alpha matte prediction stage.

The loss function of training QUN network contains two parts, identity loss and consistence loss. Identity loss forces the output of QUN not to change much from the original input,
\begin{equation}
\begin{aligned}
\label{identity_loss}
\mathcal{L}_{identity}= |Q(x)-x|_1+|Q(x')-x'|_1\,,
\end{aligned}
\end{equation}
where $Q(\cdot)$ represent the quality unification network. $x$ denotes the concatenation of the input image and the accurate mask, $x'$ denotes the concatenation of the input image and the inaccurate mask. The second part is consistence loss. Consistence loss forces the output of QUN corresponding to accurate mask and inaccurate mask to be close.

\begin{equation}
\begin{aligned}
\label{consist_loss}
\mathcal{L}_{consist}=|Q(x)-Q(x')|\,.
\end{aligned}
\end{equation}

Thus, the loss function of training QUN is the weighted sum of identity loss and consistence loss,

\begin{equation}
\begin{aligned}
\label{QUN_loss}
\mathcal{L}_{QUN}= \lambda_1\mathcal{L}_{identity}+ \lambda_2\mathcal{L}_{consist}\,.
\end{aligned}
\end{equation}

During the training, we set $\lambda_1=0.25$ and  $\lambda_2=0.5$.

In Figure~\ref{fig:demo_mqun}, we illustrate the results of QUN. Fine mask (Figure~\ref{fig:demo_mqun}(a)) and coarse mask (Figure~\ref{fig:demo_mqun}(b)) are unified by QUN to Figure~\ref{fig:demo_mqun}(d) and (e) respectively. The difference maps are also calculated. We can observe that the unified high quality mask become relatively coarser and low quality mask becomes relatively finer. As a result, the unified masks are much closer to each other than the original fine and coarse masks.

\subsection{Matting Refinement Network}
Matting refinement network (MRN) aims to predict accurate alpha matte. Therefore, we train MRN at a higher resolution ($768*640$ in all experiments). Note that the coarse mask from MPN and QUN is at low resolution ($192\times 160$). The coarse mask is integrated to MRN as external input feature maps, where the input is downscaled $4$ times after several convolution operations. The output of MRN are 4-channel maps, including three foreground RGB channels and one alpha matte channel. Predicting the foreground RGB channels coupled with alpha matte is able to increase the robustness, which plays a similar role of the compositional loss used in~\cite{xu2017deep,chen2018semantic}. The loss function we used to train MRN is $L_1$ loss,
\begin{equation}
\begin{aligned}
\label{eq:hrn_loss}
\mathcal{L}_{MRN}=\lambda_H|{RGB}_p-{RGB}_g|_1+(1-\lambda_H)|\alpha_p-\alpha_g|_1\,,
\end{aligned}
\end{equation}
where ${RGB}_p$ and ${RGB}_g$ denote the predicted RGB foreground channels and ground truth foreground channels respectively. $\alpha_p$ and $\alpha_g$ denote the predicted alpha matte and ground truth alpha matte respectively. We set $\lambda_H=0.5$ in experiments.

\begin{figure}[t]
  \centering
  \resizebox{1.00\linewidth}{!}{
   \includegraphics{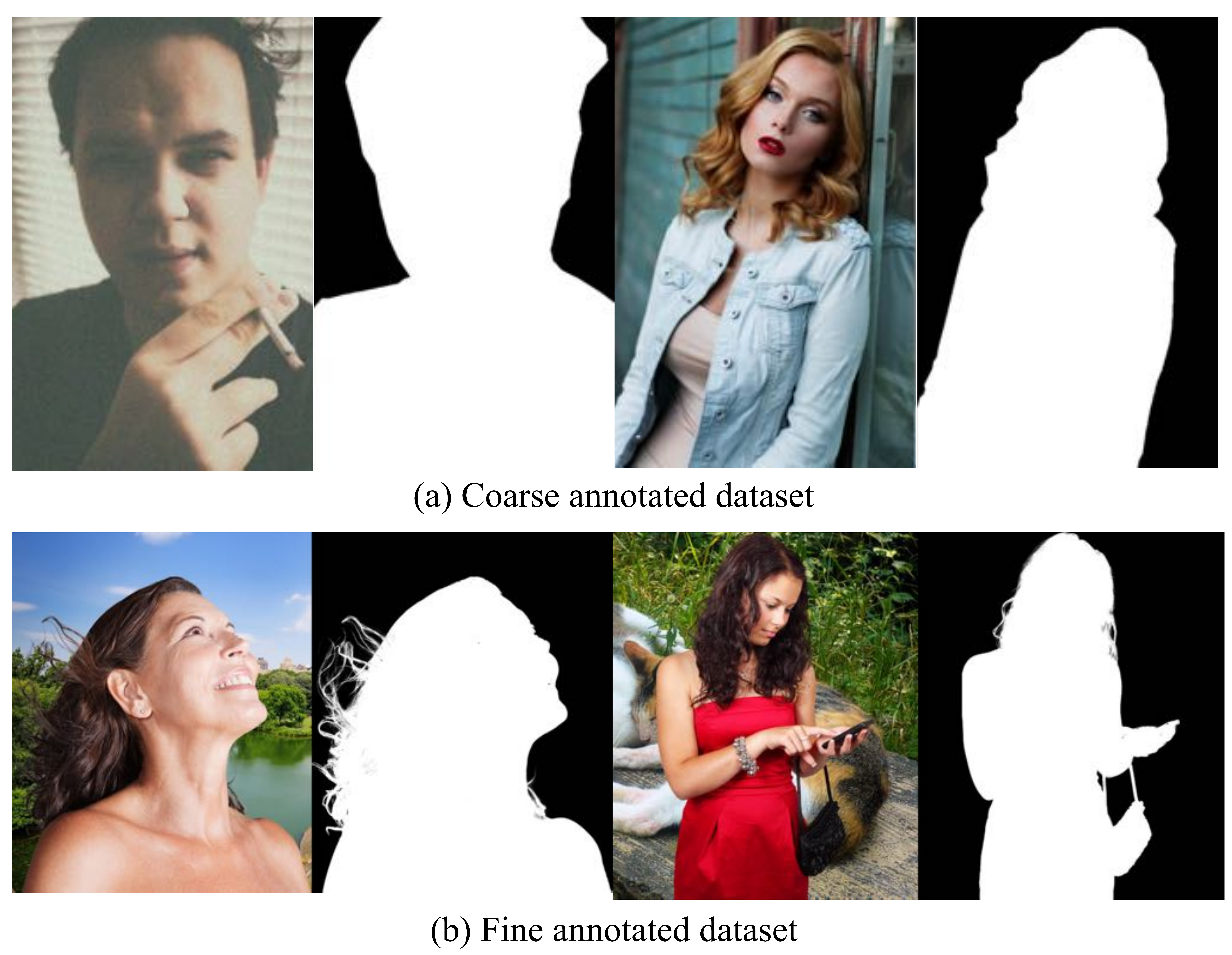} }
  \caption{Input images and the corresponding annotations in our dataset. Our dataset consists of both coarse annotated images (a) and fine annotated images (b).}
  \label{fig: dataset}
\end{figure}

\subsection{Implementation details}

We implement our method with Tensorflow~\cite{abadi2016tensorflow} framework. We perform training for our three networks sequentially. Before feeding into the mask prediction network, we conduct a down-sampling operation on images at $192\times160$ resolution, including both fine and coarse annotated data. Flipping is performed randomly on each training pair. We first train the mask prediction network for 20 epochs and fix the parameters. Then we concatenate the low resolution image and the output foreground mask as input to train quality unification network. When training QUN, random filters(filter size set as 3 or 5), binarization and morphology operations(dilate and erode) are exerted to fine annotated data to generate paired high and low quality mask data. After training quality unification network, all parameters are fixed. We finally train the matting refinement network with only the fine annotated data. The entire data pairs (image, alpha matte) are randomly cropped to $768\times640$. The learning rate for training all networks is $1e-3$. MPN and QUN are trained using batch size $16$ and MRN is trained using batch size $1$, as MRN is trained using only high resolution data.

When testing, a feed-forward pass of our pipeline is performed to output the alpha matte prediction with only the image as input. The average testing time on multiple 800$\times$800 images is 0.08 seconds.

\begin{table}[t]
  \centering\scriptsize
    \caption{The configurations of human matting datasets.}
  \resizebox{1\linewidth}{!}{
  \begin{tabular}{ccccc}
    \hline
    \multirow{2}{*}{Dataset} &
    \multicolumn{2}{c}{Train Set} & \multicolumn{2}{c}{Test Set} \\
     &  Human & image & Human & image \\
    \hline
    Shen \etal~\cite{shen2016deep} & 1700 & 1700 & 300 & 300 \\
    TrimapDIM~\cite{xu2017deep} & 202 & 20200 & 11 & 220 \\
    SHM~\cite{chen2018semantic} & 34493 & 34493 & 1020 & 1020 \\
    Ours(coarse) & 10597 & 105970 &  \multirow{2}{*}{125 (+11)} & \multirow{2}{*}{1360} \\
    Ours(fine) & 9324(+202) & 95260 &  &  \\
    \hline
    \end{tabular}}
  \label{tab:dataset}
\end{table}

\section{Human matting dataset}

A main challenge for human matting is the lack of data. Xu \etal~\cite{xu2017deep} proposed a general matting dataset by compositing foreground objects from natural images to differents backgrounds, which has been widely used in the following matting works\cite{Cai_2019_ICCV,lutz2018alphagan,Zhang_2019_CVPR}. However, the diversity of human images is severely limited, including only 202 human images in training set and 11 human images in testing set. For human matting dataset, Shen \etal~\cite{shen2016deep} collected a portrait dataset with 2000 images, it assumes that the upper body appears at similar positions in human images and the images are annotated by Closed From~\cite{levin2007closed}, KNN~\cite{chen2013knn} methods, which can be inevitably biased. Although a large human fashion dataset is created by \cite{chen2018semantic} for matting, it is only for commercial use. To this end, we create a human matting dataset with high-quality for research. We carefully collected 9449 diverse human images with simple background from the Internet (i.e., white or transparent background in PNG format), each human image acquires a well annotated alpha matte after simple processing. The human images are split to training/testing set, with 9324 and 125 respectively. Following Xu \etal~\cite{xu2017deep}, we first add the human images in DIM dataset\cite{xu2017deep} into our training/testing set, forming a total of 9526 and 136 human foregrounds respectively. We then randomly sample 10 background images in MS COCO~\cite{lin2014microsoft} and Pascal VOC~\cite{everingham2010pascal} and composite the human images onto those background images. During composition, we ensure that the background images are not containing humans.

\begin{figure*}[t!]
  \centering
  \resizebox{1.00\linewidth}{!}{
   \includegraphics{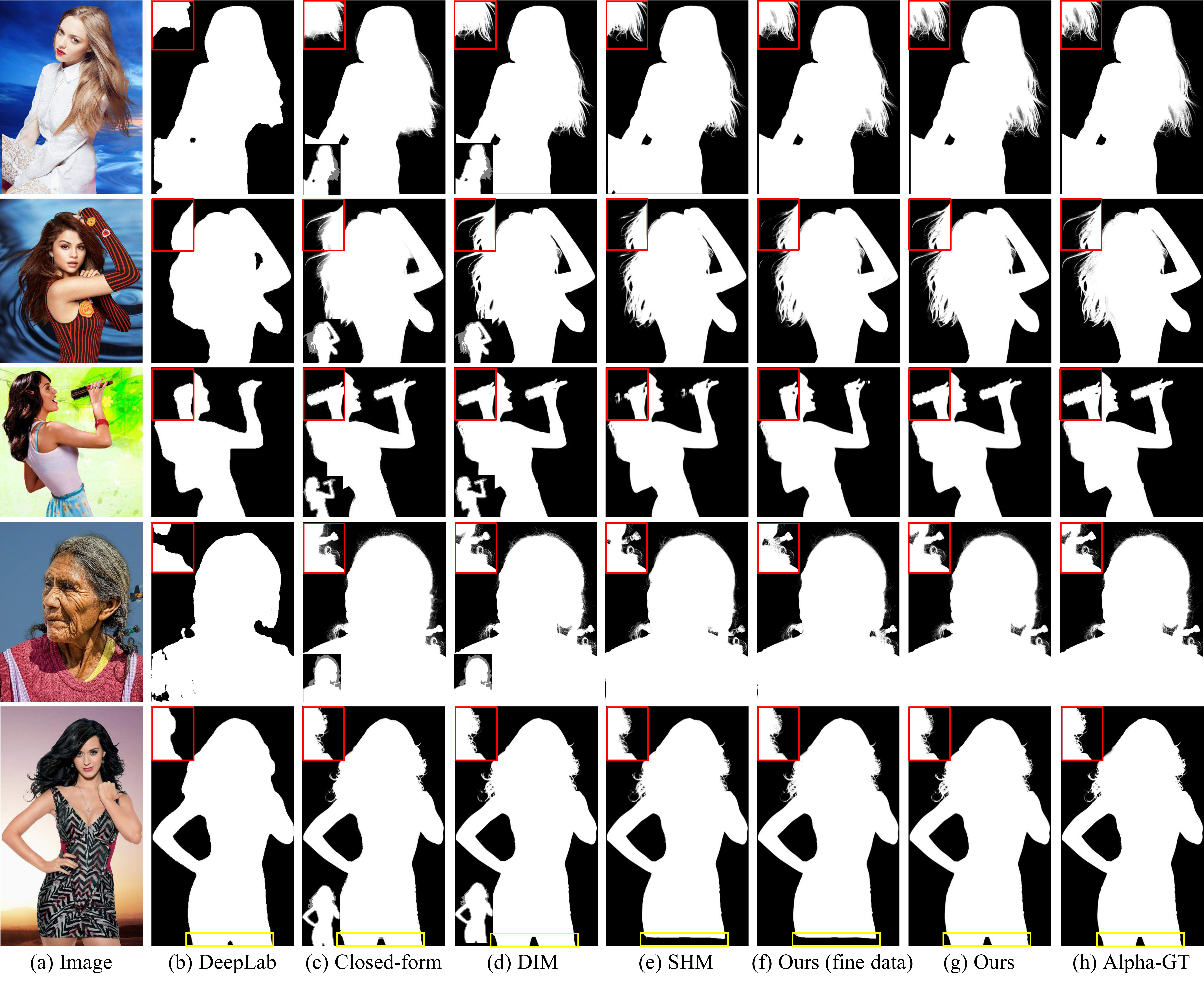} }
  \caption{The qualitative comparison on our proposed dataset. The first column and the last column show the input image and the ground truth alpha matte, and the rest columns present the estimation results by DeepLab~\cite{chen2017rethinking}, Closed-form matting~\cite{levin2007closed}, DIM~\cite{xu2017deep}, SHM~\cite{chen2018semantic}, our method trained using fine annotated data only and our method trained using hybrid annotated data.}
  \label{fig:quality_cmp}
\end{figure*}

Another issue should be addressed for human matting dataset is the quality of annotations. Image matting task requires user designated annotations for objects, i.e., the high quality alpha matte. Besides, the user interaction methods require carefully prepared trimaps and scribbles as constraints, which is labor intensive and less scalable. Method without user provided trimaps is to predict the alpha matte by first generating implicit trimaps for further guidance, thus lead to some artifacts as well as losing some semantics for complex structures. We integrate the coarse annotation data to tackle this problem as they are much easier to obtain. We collect another 10597 human data from~\cite{wu2014early} and Supervisely Person Dataset, and follow the above setup to generate 105970 image with coarse annotations.

Table~\ref{tab:dataset} shows the configuration of the existing human matting dataset. Our dataset consists of both fine and coarse annotated data, with nearly the same amount. Compared with user interactive methods~\cite{shen2016deep,xu2017deep}, our dataset covers diverse high quality human images, making it more robust for human matting models. Although sacrifice the number of high quality annotations than automatic method~\cite{chen2018semantic}, we introduce coarse annotated data to enhance the capacity for extracting both semantic and matting details at a lower cost. The data for both annotations are shown in Figure~\ref{fig: dataset}.

\section{Experiments}
\label{sec: Experiments}

\subsection{Evaluation results.}

\paragraph{Evaluation metrics.} We adopt four widely used metrics for matting evaluations following the previous works~\cite{xu2017deep,chen2018semantic}. The metrics are MSE (mean square error), SAD (sum of the absolution difference), the gradient error and the connectivity error. The gradient error and connectivity error proposed in~\cite{rhemann2009perceptually} are used to reflect the human perception towards visual quality of the alpha matte. Lower values of these metrics correspond to better estimated alpha matte.  We normalize the estimated alpha matte and true alpha matte to [0, 1] to calculate these evaluation metrics. Since no trimap is required, we calculate over the entire images and average by the pixel number.

\paragraph{Baselines.} We select the most typical method from semantic segmentation methods, traditional matting methods, user interactive methods and automatic methods respectively as our baselines. These methods are DeepLab~\cite{chen2017rethinking}, Closed-form matting~\cite{levin2007closed}, DIM~\cite{aksoy2017designing} and SHM~\cite{chen2018semantic}. Note that the Closed-form matting and DIM need extra trimap as input. DIM and SHM can only be trained using the fine annotated data. DeepLab and the proposed method are trained using the proposed hybrid annotated dataset.

\begin{table}[t]
  \centering\scriptsize
    \caption{The quantitative results.}
  \resizebox{1\linewidth}{!}{
  \begin{tabular}{lcccc}
    \hline
    Method & SAD & MSE & Gradient & Connectivity \\
    \hline
    DeepLab ~\cite{chen2017rethinking} & 0.028 & 0.023 & 0.012 & 0.028 \\
    Trimap+CF ~\cite{levin2007closed} & 0.0083 & 0.0049 & 0.0035 & 0.080 \\
    Trimap+DIM~\cite{xu2017deep} & \textbf{0.0045} & \textbf{0.0017} & \textbf{0.0013} & \textbf{0.0043} \\
    SHM~\cite{chen2018semantic} & 0.011 & 0.0078 & 0.0032 & 0.011 \\
    \hline
    ours(w/o coarse data) & 0.0099 & 0.0067 & 0.0029 & 0.0095 \\
    ours(w/o QUN) & 0.0076 & 0.0042 & 0.0024 & 0.0072 \\
    \hline
    ours & \textbf{0.0058} & \textbf{0.0026} & \textbf{0.0016} & \textbf{0.0054} \\
    \hline
    \end{tabular}}
  \label{tab:quantitative}
\end{table}

\paragraph{Performance comparison.}
In Table~\ref{tab:quantitative}, we list the quantitative results over 1360 testing images. The semantic segmentation method DeepLab~\cite{chen2017rethinking} only predict coarse mask and lack fine details (Figure~\ref{fig:quality_cmp}(b)), resulting in the worst quantitative metrics. SHM~\cite{chen2018semantic} does not perform well as the volume of our high quality training dataset is limited, and fails to predict accurate semantic information for some images (Figure~\ref{fig:quality_cmp}(d)). In contrast, the interactive method close-form matting~\cite{levin2007closed} and DIM~\cite{xu2017deep} performs well, benefiting from the input semantic information provided by trimaps. These two methods only need to estimate the uncertain part in trimaps. The proposed method using hybrid training dataset outperforms most methods and is comparable with state-of-the-art methods. DIM~\cite{xu2017deep} is slightly better than the proposed method. Note that the proposed method only take in input images, DIM requires high informative trimaps as extra input. Even though, the visual quality of the proposed method (Figure~\ref{fig:quality_cmp}(g)) and DIM (Figure~\ref{fig:quality_cmp}(d)) looks very close.



\begin{figure}[ht]
  \centering
  \resizebox{1.00\linewidth}{!}{
   \includegraphics{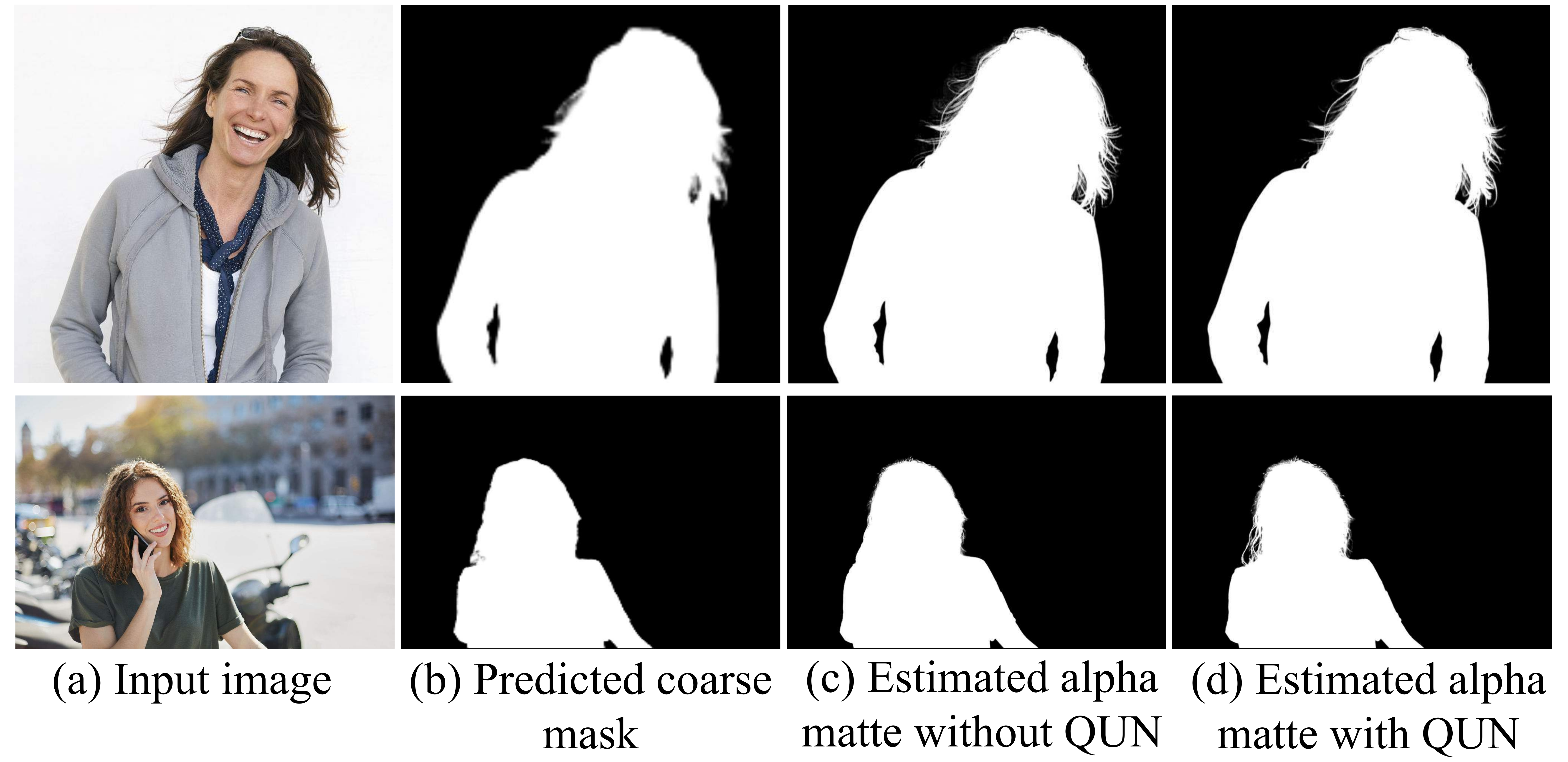} }
  \caption{Self-comparisons. Without quality unification network (QUN), the quality of coarse mask sent to the matting refinement network (MRN) may vary significantly. When the coarse mask is relatively accurate, MRN predicts alpha matte well. When the coarse mask lacks most hair details, the estimated alpha matte is accurate. Equipped with QUN, the mask quality is unified before feeding into MRN. The estimated alpha matte is more consistent against different kinds of coarse masks.}
  \label{fig:mid_figure}
\end{figure}

\begin{figure}
  \centering
  \resizebox{1.00\linewidth}{!}{
   \includegraphics{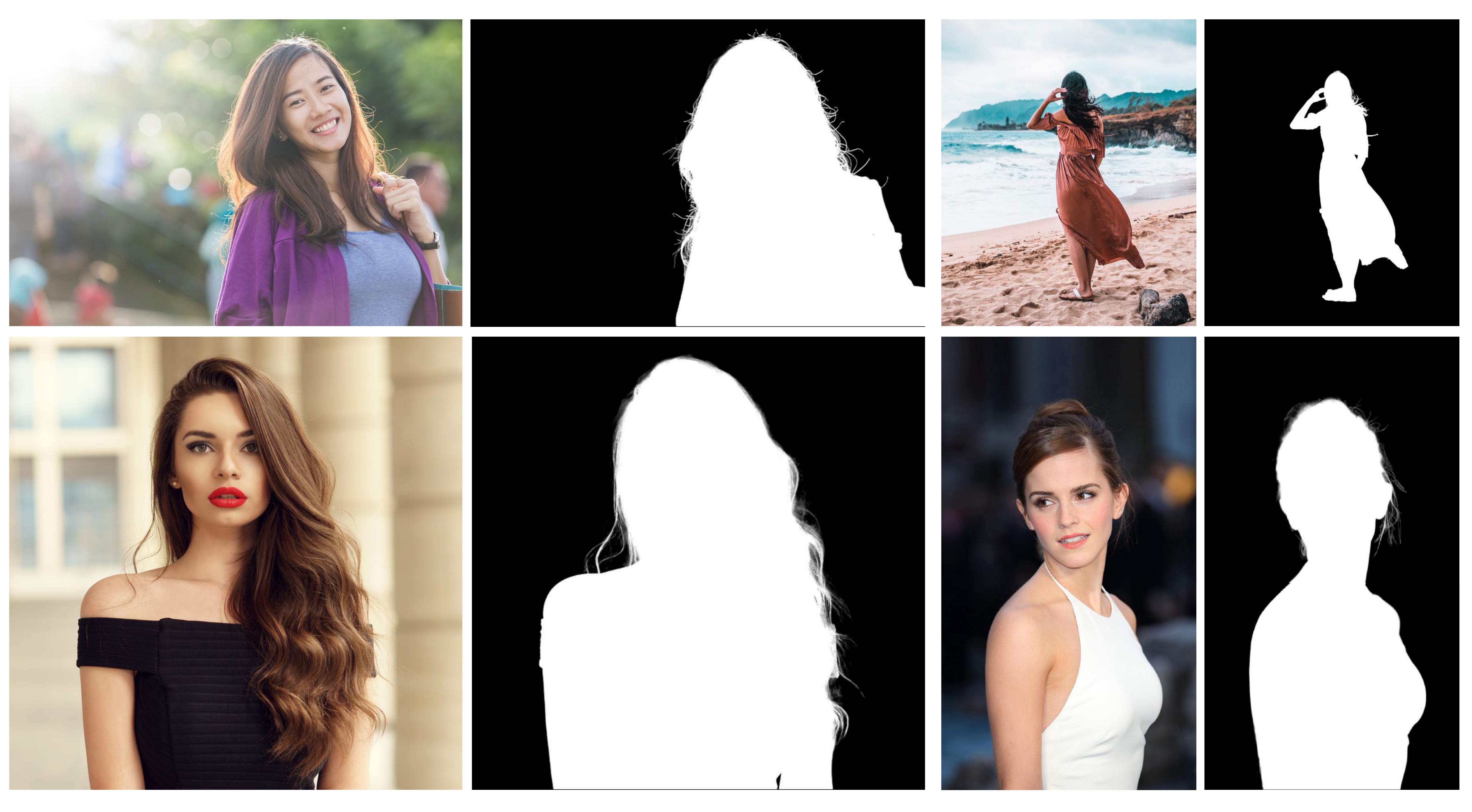} }
  \caption{Real image matting results. The collected coarse annotated dataset enriches our dataset significantly and enables the proposed method to capture the semantic information well and predicts accurate alpha matte for different kinds of input images.}
  \label{fig:real_image}
\end{figure}

\paragraph{Self-comparisons.} Our method can achieve high quality alpha matte estimation by incorporating coarse annotated human data. Coarse annotated data promote the proposed network to estimate semantic information accurately. To verify the importance of the these data, we separately train the same network with fine annotated dataset only. The quantitative results are listed in Table~\ref{tab:quantitative}. Without using the coarse data, the performance is obviously worse. From Figure~\ref{fig:quality_cmp}(f) and (g), we can also observe that method trained only with fine annotated data suffers from inaccurate semantic estimation and presents incomplete alpha matte.

The mask quality unification network make it possible for the final matting refinement network to adapt to different kinds of coarse mask input. Without QUN, inputs to the matting refinement network may vary significantly, which is hard to deal with at inference stage. We list the quantitative metrics without QUN being used in Table~\ref{tab:dataset}. Both fine and coarse annotated dataset are used in this comparison. The results are obviously worse when QUN is removed. For a better visual comparison, we display the results in Figure~\ref{fig:mid_figure}. The predicted alpha matte is fine if the coarse mask is relatively accurate. When the coarse mask lacks most hair details, the estimated alpha matte is not good. With QUN, the mask quality is unified before feeding into MRN. The estimated alpha matte is more accurate and robust to different kinds of coarse masks.

\begin{figure}[t!]
  \centering
  \resizebox{1.00\linewidth}{!}{
   \includegraphics{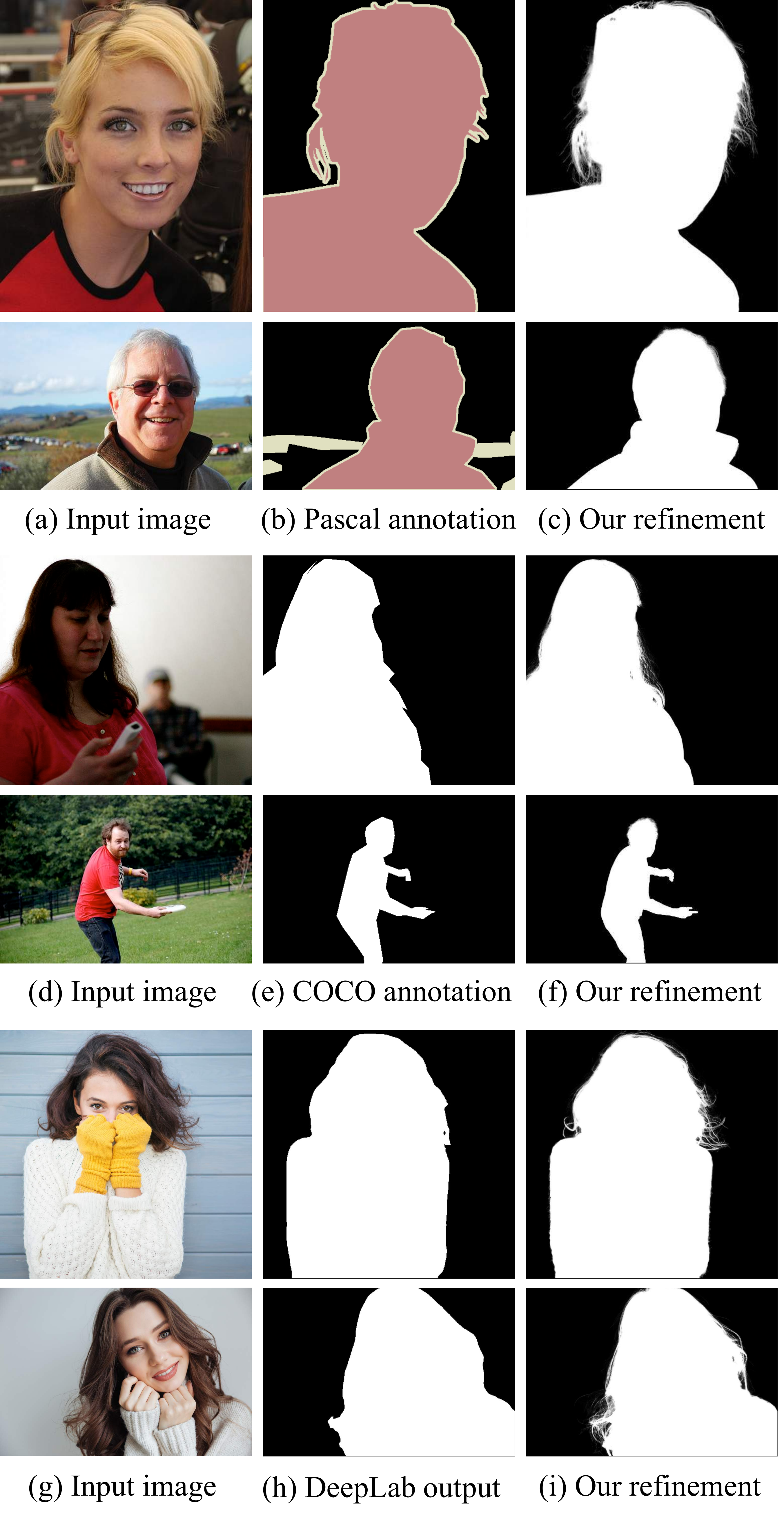} }
  \caption{Using the proposed method to refine coarse human mask from public dataset annotations or semantic segmentation methods. Feed the coarse human mask from Pascal (b) or Coco (e) dataset annotation or DeepLab (h) to our quality unification network, and then use the matting refinement network to generate the accurate human alpha matte.}
  \label{fig:application}
\end{figure}

\subsection{Applying to real images}
We further apply the proposed method to real images from the Internet. Matting on real images is challenging as the foreground is smoothly fused with the background. In Figure~\ref{fig:real_image}, we display our testing results on real images. Benefiting from the sufficient training on our hybrid dataset, the proposed method captures the semantic information very well for different kinds of input images and predicts accurate alpha matte at a detailed level.


\section{Applications}
The mask prediction network in the proposed method aims to capture coarse semantic information requiring by the subsequent networks. The semantic mask from this network can be coarse or accurate. The following quality unification network will unify the mask quality for the final matting refinement network. Therefore, if the semantic mask is arrange in some way, the proposed method is still able to work seamlessly and generate accurate alpha matte.

Thus we can apply our framework to refine coarse annotated public dataset, such as the PASCAL~\cite{pascal-voc-2007} (Figure~\ref{fig:application}(a-c)) and COCO dataset~\cite{lin2014microsoft} (Figure~\ref{fig:application}(d-f)). The annotated human mask are resized and used as input for our QUN and MRN. Even though the annotations are not accurate, especially the annotations from COCO dataset, the proposed method manages to generate accurate refinement results.

We can also use the proposed method to refine semantic segmentation methods (Figure~\ref{fig:application}(g-i)). Semantic segmentation methods are usually trained on coarse annotated public dataset, and the output mask is not precise. We feed the coarse mask obtained from DeepLab~\cite{chen2017rethinking} to our QUN and MRN. The proposed method generates surprisingly good alpha matte. Details that are missing from the coarse mask are well recovered, even for the very detailed hair parts.


\section{Conclusion}
\label{sec: Conclusion}
In this paper, we propose to use coarse annotated data coupled with fine annotated data to enhance the performance of end-to-end semantic human matting. We propose to use MPN to estimate coarse semantic masks using the hybrid annotated dataset, and then use QUN to unify the quality of the coarse masks. The unified mask and the input images are fed into MRN to predict the final alpha matte. The collected coarse annotated dataset enriches our dataset significantly, and makes it possible to generate high quality alpha matte for real images. Experimental results show that the proposed method performs comparably against state-of-the-art methods. In addition, the proposed method can be used for refining coarse annotated public dataset, as well as semantic segmentation methods, which potentially brings a new method to annotate high quality human data with much less effort.
\clearpage

{\small
\bibliographystyle{ieee_fullname}
\bibliography{egbib}
}

\end{document}